\documentclass[final,1p,times]{elsarticle}

\usepackage{amssymb}

\usepackage[colorlinks,
            linkcolor=blue,
            anchorcolor=blue,
            citecolor=blue]{hyperref}
\biboptions{numbers,sort&compress}
\journal{Image and Vision Computing}

\usepackage{url}
\usepackage{color}
\usepackage{indentfirst}
\usepackage{graphicx}     
\usepackage{amsmath}
\usepackage{amssymb}
\usepackage{algorithm} 
\usepackage{algpseudocode} 
\usepackage{float}
\usepackage{subcaption}
\usepackage{multirow}
\usepackage{epsfig}
\begin{document}

\begin{frontmatter}



\title{Conditional Generative Data-free Knowledge Distillation}


\author[1]{Xinyi Yu}
\ead{yuxy@zjut.edu.cn}
\author[1]{Ling Yan}
\ead{lingyan@zjut.edu.cn}
\author[1]{Yang Yang}
\ead{yangyang98@zjut.edu.cn}
\author[1]{Libo Zhou}
\ead{libozhou@zjut.edu.cn}
\author[1]{Linlin Ou\corref{mycorrespondingauthor}}
\cortext[mycorrespondingauthor]{Corresponding author}
\ead{linlinou@zjut.edu.cn}

\address[1]{{College of Information and Engineering, Zhejiang University of Technology},
            {Hangzhou, Zhejiang},
            {China}}

\begin{abstract}
Knowledge distillation has made remarkable achievements in model compression. However, most existing methods require the original training data, which is usually unavailable due to privacy and security issues. This paper proposes a conditional generative data-free knowledge distillation (CGDD) framework for training lightweight networks without real data. This framework realizes efficient knowledge distillation based on conditional image generation. Specifically, we treat the preset labels as ground truth to train a semi-supervised conditional generator. The trained generator can produce specified classes of training images. During training, we force the student model to extract the hidden knowledge in teacher feature maps, which provide crucial cues to the learning process. Meanwhile, we construct an adversarial training framework to promote distillation performance. The framework will help the student model to explore larger data space. To demonstrate the effectiveness of the proposed method, we conduct extensive experiments on different datasets. Compared with other data-free works, our method obtains state-of-the-art results on CIFAR100, Caltech101, and different versions of ImageNet datasets. The codes will be released.
\end{abstract}




\end{frontmatter}


\section{Introduction}\label{sec1}

Deep learning has achieved great performance in various computer vision applications \cite{bib1, bib2, bib3, bib4, bib5}. The superior performance came with over-parameterized models. However, ever-growing model parameters and computational costs suppress the application of high-performance models on cloud and edge devices. Various effective techniques\cite{bib6, bib7, bib8} have been proposed to compress and speed up the heavy models to address this problem. In this case, knowledge distillation (KD) \cite{bib9, bib10, bib11, bib12, bib13, bib14, bib15, bib16} also draws extensive attention benefit from good model compression effect, and it has extensive applications in Computer Vision (CV) \cite{bib9, bib10, bib11}, Natural Language Processing (NLP) \cite {bib12, bib13, bib14}, and speech recognition \cite{bib15, bib16}.
\par

Specifically, KD is the method to help the small student model training under the guidance of the teacher network. In deep models, the pre-trained teacher networks contain ample information about the original training data and target tasks, called knowledge. KD aims to transfer the knowledge to student models and help them train. For the convolutional neural networks, Hinton et al. \cite{bib9} first attempted to distill the knowledge to student networks by minimizing the discrepancy between network logits.
Further works utilize the latent information in intermediate layers to improve the distillation performance. FitNets \cite{bib10} utilizes the outputs (logits) and intermediate features together to trains the student models. Zagoruyko et al. \cite{bib11} promoted student performance by forcing it to imitate teacher attention maps.\par 
 
KD can efficiently compress cumbersome models when original or alternative data is available. However, training datasets for given models are often unavailable due to privacy and security limitations, such as federated learning tasks \cite{bib17, bib18, bib19}. In addition, collecting alternative data is onerous, while irrelevant data will drastically deteriorate the performance of student networks. An effective way that tackles these questions is to reconstruct data as training samples. However, although the pre-trained teacher model contains rich knowledge of original training data, it is difficult to exploit the latent knowledge and reconstruct meaningful samples. Fortunately, generative adversarial networks (GANs \cite{bib20}) offer a powerful capability to exploit data distribution and craft images. Recently, several data-free (zero-shot) knowledge distillation methods \cite{bib21, bib22,bib23,bib24}that exploit generative adversarial networks have been proposed for compressing heavy models when real data is unavailable. And it can be employed on many extended tasks, such as the federated learning of medical data. However, due to the absence of conventional generation techniques such as GANs or VAE \cite{bib25} cannot be directly applied to data-free KD. To this end, Chen et al. \cite{bib22} trained a generator to create one-hot-like images and help the model distillation. These images will highly activate the neurons in teacher models. Fang et al. \cite{bib23} presented an adversarial distillation mechanism to move generated samples toward the areas where the current student has not been well trained. Luo et al. \cite{bib24} leveraged the intrinsic batch normalization statistics in the teacher model to guide the training of generators.\par

\begin{figure}
	\centering
	\includegraphics[width=1.0\textwidth]{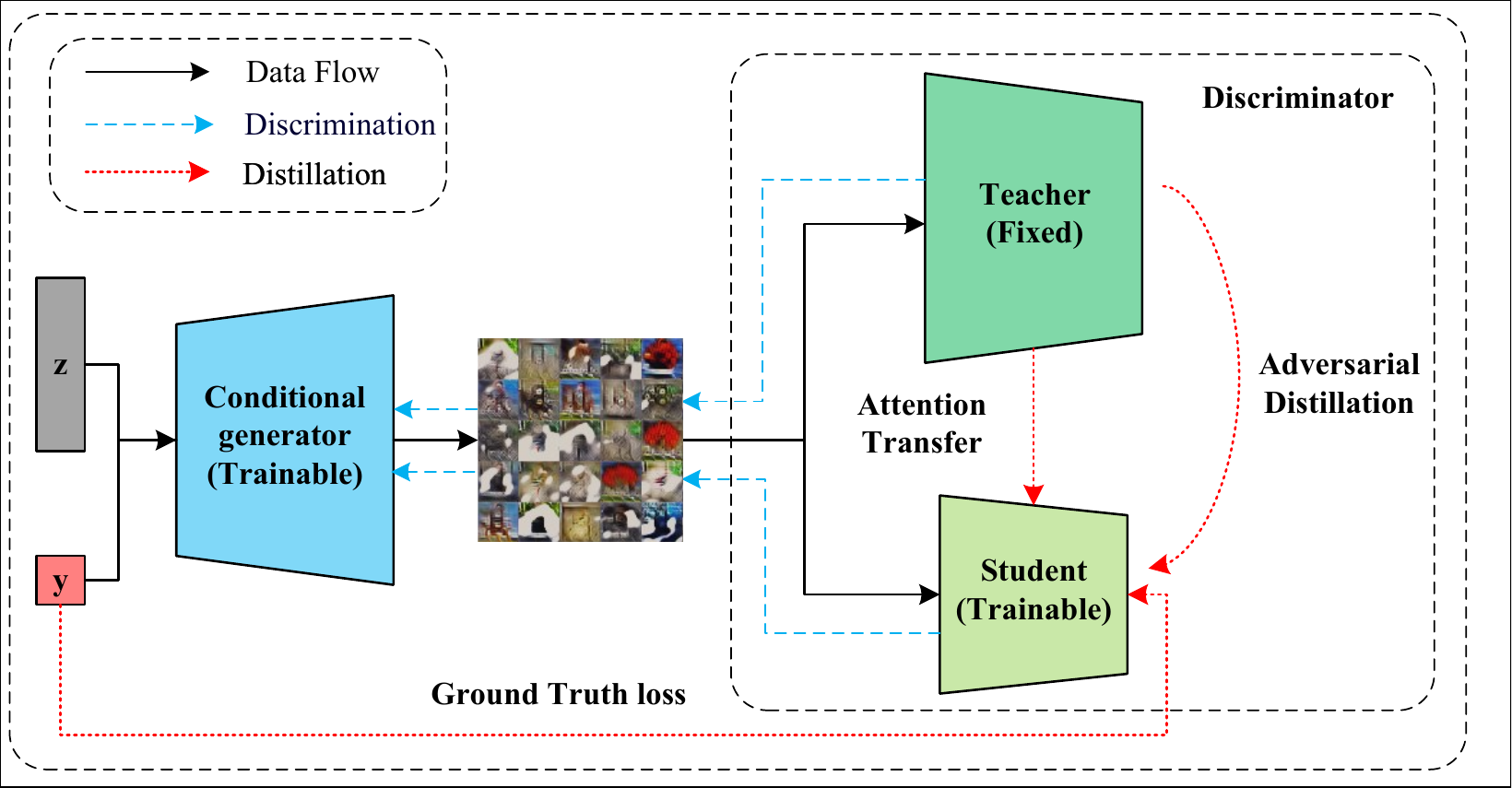}
	\caption{The framework of the conditional generative data-free knowledge distillation method. In the adversarial training framework, the generator is guided to craft meaningful training data, and the student and teacher networks are combined as a discriminator. The student is optimized under the supervision of a teacher.}
	\label{fig1}
\end{figure}
Although there are several data-free knowledge distillation methods, a performance gap still exists between the trained teacher and student models, while the data-driven KD can achieve even better performance than teacher models. In addition, the trained generator only produces class-balanced synthetic images that may be incompatible with some unique settings. Such as federated learning tasks which concern the class-imbalanced issue. \par

In this paper, a conditional generative data-free distillation framework is designed to attain efficient image generation and distillation, as illustrated in Fig.\ref{fig1}. Specifically, we introduce preset labels as ground truth to build a semi-supervised conditional generator. With this supervision, we can adapt label distribution to match different datasets and avoid heterogeneity issues. Since preset labels will determine image classes, the student model can be optimized under the guidance of ground truth (preset labels). Besides distilling the outputs (i.e., logits), we also utilize the feature information (e.g., attention maps) that provides more precise and efficient guides to the learning process. Precisely, we force the student network to learn the feature information in the teacher model. It validates the importance of feature information and significantly improves distillation performance under data-free settings. \par

To demonstrate the efficacy of our method, we perform extensive experiments on several datasets and conduct a series of ablation experiments to analyze different designs. The main contributions of this work are summarized as follows: \par

$\bullet$ We propose a conditional generative data-free distillation framework, which consistently produces meaningful training samples and attains efficient knowledge distillation.\par

$\bullet$ Our method trains a semi-supervised conditional generator by introducing preset labels as ground truth. The semi-supervised training process can address the class distribution issue by label sampling. \par

$\bullet$ We force the student model to learn teacher knowledge and consistently explore unseen data space in the adversarial training process. It can fully extract the feature information and improves student performance.\par

$\bullet$ Experiments on MNIST, CIFAR, Caltech101, and part of ImageNet datasets are conducted to demonstrate the superiority of our method. We obtain state-of-the-art on most datasets.\par

\section{Related Work}\label{sec2}

In this section, we briefly review existing works on related topics. \par

\subsection{Data-driven Knowledge Distillation}

Knowledge distillation is a general model compression technique. In KD, we can transfer knowledge by minimizing network logit difference. However, teacher logits usually have a high probability in the correct class and close to zero in other classes. It cannot provide more information than the one-hot label. To this end, Hinton et al. \cite{bib9} introduced SoftMax temperature to soften model output and extract more relationship information between different classes. Furthermore, Furlanello et al. \cite{bib26} trained and integrated a set of student networks as the final student model. Mirzadeh et al. \cite{bib27} introduced a middle-sized model (teacher assistant) as a bridge to help the student network training.

Besides distilling the logits, transferring knowledge to the student can be achieved by utilizing the feature information in hidden layers. Feature knowledge contains precise and efficient cues of the training process. In this case, Romero et al. \cite{bib10} trained student networks using both logits and intermediate features. Zagoruyko et al. \cite{bib11} transformed latent knowledge into attention maps and forced student networks to mimic the attention maps of the teacher. Yim et al. \cite{bib28} introduced the FSP (Flow of Solution Procedure) matrix to learn the relationship between inputs and outputs. Park et al. \cite{bib29} proposed a relational knowledge distillation method. They transferred the mutual relations of data according to the distance-wise loss and angle-wise loss. Bai et al. \cite{bib30} designed a novel layer-wise cross distillation method and realize few-shot distillation. Li et al. \cite{bib31} trained a portable objective detection model based on the supervision from proposal features. Liu et al. \cite{bib32} presented a dense prediction distillation method based on holistic knowledge and pairwise similarity. 
Some recent works \cite{bib33, bib34} also explored the structure effect to model distillation. Moreover, combining the knowledge distillation with other techniques (e.g., NAS \cite{bib34}, RL \cite{bib35}, or GNN \cite{bib36}) can further compress and accelerate the heavy models. For other related works, \cite{bib37} reviewed and looked at the development of knowledge distillation.

Although these data-driven KD methods have achieved outstanding performance in many applications, there are many challenges when real data is unavailable. Exploring more efficient knowledge representation and utilization techniques for data-free tasks is necessary.

\subsection{Data-free Knowledge Distillation}

Since real data is inaccessible due to data privacy and security issues, many researchers made efforts on data-free compression techniques, including data-free pruning \cite{bib38, bib39}, data-free quantization \cite{bib40, bib41,bib42}, and data-free knowledge distillation \cite{bib21,bib22,bib23,bib24,bib43,bib44,bib45,bib46}. This paper focuses on the data-free KD and trains lightweight student networks through transferring teacher knowledge.

In data-free KD, Lopes et al. \cite{bib21} attempted to reconstruct original data from metadata and utilized the synthetic data for training student networks. However, it still requires metadata extracted from real data. Nayak et al. \cite{bib43} modeled the SoftMax space as a Dirichlet distribution and updated random noise images to craft Data Impressions. Chen et al. \cite{bib22} designed a novel framework (DAFL) to train a generator that matches the original data distribution. They fixed the teacher model as a discriminator and designed several novel loss functions to optimize the generator. In contrast, Fang et al. \cite{bib23} presented an adversarial distillation mechanism (DFAD). In DFAD, teacher and student jointly play the role of a discriminator to reduce prediction discrepancy, while the generator adversarially enlarges the discrepancy by creating hard samples. Yin et al. \cite{bib44} synthesized images through model inversion and needed no additional generators. They utilized the information from batch normalization statistics to improve synthetic images. For data-free distillation on large-scale datasets, Luo et al. \cite{bib24} trained one generator for each class. It gains remarkable performance and avoids mode collapse (the generator produces similar images and loses diversity) but requires vast computational resources. Meanwhile, Jin et al. \cite{bib45} used a simple fully connected network as an additional discriminator to distinguish the teacher and student models. Fang et al. \cite{bib46} improved image diversity by introducing contrastive learning.

These methods have realized data-free distillation, but the feature information which provides crucial cues to the learning process is not fully used. In this case, the student model is unable to obtain desirable performance. Moreover, the generator can use the class information and tackle heterogeneity issues through label shift. In this work, we propose a more efficient approach to attain data-free image generation and distillation.

\subsection{Generative Adversarial Networks}
To transfer knowledge from the teacher to the student network when training data is inaccessible, we adopt the framework of GAN to produce training samples. GAN contains two interacting neural networks: a generator $G$ that captures data distribution and a discriminator $D$ to distinguish the real and fake samples (generated by $G$). The two networks are trained jointly through a minimax game.
In this adversarial process, $G$ learns to generate meaningful samples that the match original distribution. To make the generator produce specified classes of images, Mirza et al. \cite{bib47} proposed the conditional GAN. It applies class conditions to both generator and discriminator networks. Furthermore, Arjovsky et al. \cite{bib48} presented Wasserstein GAN (WGAN) to overcome training instability and mode collapse issues. Similarly, Qi et al. \cite{bib49} regularized the adversarial loss through the Lipschitz regularization.

In recent years, GAN has achieved great success in many image synthesis areas \cite{bib50}. Some works attempted to improve computer vision tasks through the GAN module. To help the training of the fine-grained human action segmentation mode, Gammulle et al. \cite{bib51} developed a recurrent semi-supervised action GAN, which learns an intermediate representation. Zheng et al. \cite{bib52} enhanced the person re-identification (re-ID) model based on a generative module. The generative module learns to produce better images for conducting teacher-student distillation, while the re-id model (discriminative module) mines and extracts better subtle features. Zhang et al. \cite{bib53} designed a semi-supervised GAN for precise automatic defect detection that can utilize unlabeled data to improve segmentation results. These works apply the GAN module to enhance different deep models in a semi-supervised learning manner. Theoretically, the powerful GAN module can be applied for data-free KD. However, GAN-based methods aim to craft realistic images and require real training data. Therefore, developing an efficient data-free GAN framework is necessary when conventional GAN fails in data-free tasks. We train a semi-supervised conditional generator to produce training data by exploring the hidden original data distribution in the teacher model.

\section{Method}\label{sec3}
This section describes the proposed conditional generative data-free distillation framework. As shown in Fig.\ref{fig1}, the framework contains three networks: a pre-trained teacher network $T$, a smaller student network $S$, and a generator $G$. During training, a set of random noise $z=\left\{z^1,z^2,...,z^n\right\}$ is inputted into $G$ to produce training samples $G(z)$, where $n$ is the number of samples, $z^i\in R^m$ is sampled from the normal distribution $p_z$.

\subsection{Generative Loss Design}\label{sec3.1}
In previous data-free works, the trained generators only produced images randomly. It has no clear optimization objective for the generator and classifiers (i.e., teacher and student models). However, class information is crucial for distillation and data generation, so we introduce preset labels as ground truth and build a conditional generator in this work. Under the supervision of preset labels, the trained generator can produce images of specified classes and meet different requirements (e.g., class-imbalanced datasets). To be specific, we uniformly sample a set of preset labels $\left\{ y^i\right \}$ from $\left\{0,1,...,c-1\right\}$ as conditional information, where $c$ is the number of classes. The class matching loss can be defined as:

\begin{equation}
	L_{CM}=\frac{1}{n}\sum_{i=1}^{n}H\left({\rm softmax} \left(l_T^i\right),y^i\right)\label{eq1}
\end{equation}
where $y^i$ is the preset label of $i$-th image, and $H(\cdot)$ denotes cross-entropy loss. $l_T^i=T\left(G\left(z^i\mid y^i\right)\right)$ is the logits of teacher network. $ G(z\mid y) $ is the synthetic image of class $y$. This loss function calculates the cross-entropy between teacher logits and preset labels. In order to guide the generator training, we hope the teacher model can produce the same predictions as preset labels. For class-imbalanced tasks, we will sample preset labels according to label distribution. With the class matching loss, the generator has a clear optimization objective. And we convert the generator training into a semi-supervised learning process.

In practice, we do not only optimize $ L_{CM} $ when training the generator. To measure the class balance of generated images, we introduce information entropy loss and formulate it as:
\begin{equation}
	L_{ie}=-H_{info}\left(\bar{p}\right)=\frac{1}{c}\sum_{i=0}^{c-1}\bar{p}_i\log \left(\bar{p}_i\right)\label{eq2}
\end{equation}
where $\bar{p}_i=\frac{1}{n}\sum_{j}l_T^j$ is the average logit at class $ i $, $H_{info}(\cdot)$ is information entropy. $ L_{ie} $ forces the generator to produce a set of balanced samples, and it takes the minimum when $ \bar{p}_i $ equals $1/c$. Actually, this loss complements the class matching loss in Eq.\ref{eq1} and can be utilized together to promote the image class balance. Noting that, the information entropy loss will be adjusted for class-imbalanced tasks.

Besides information entropy, teacher output is critical information of the training data. In classification tasks, high-performance models produce one-hot-like vectors for input images. If synthetic samples follow the same distribution as original data, teacher models will also output one-hot-like vectors. In other words, the input image is matching original data distribution when the teacher outputs a one-hot-like vector. To this end, we compute the cross-entropy between predicted labels and teacher logits and formulate the one-hot loss as:
\begin{equation}
	L_{oh}=\frac{1}{n}\sum_{i=1}^{n}H\left(l_T^i,y_T^i\right) \label{eq3}
\end{equation}
in which $y_T^i=\arg \max \limits_{j}\left(l_T^i\right)_j$ is the prediction of the teacher model. With the supervision of one-hot loss, we guide the generator to craft images that match the original data contribution. The one-hot loss can also be used with the class matching loss: $ L_{CM} $ requires the synthetic image to be a specified class, while $ L_{ie} $ enables the images to have more prominent features of a specific class. Since information entropy loss and one-hot loss have no clear optimization objective, we combine them as the unsupervised loss:
\begin{equation}
	L_{US}=L_{oh}+\lambda_{ie}L_{ie} \label{eq4}
\end{equation}
where $\lambda_{ie}$ is scale weight for balancing different components.

Loss functions in Eq.\ref{eq1} and Eq.\ref{eq4} utilize the knowledge from the teacher model. In data-free distillation, however, the student network is trained together with the generator. Naturally, student information is beneficial for training a generator compatible with the student. So we introduce discrepancy estimation loss to measure the gap between student and teacher predicted results. In this case, the information from the student can be used to train the generator. Moreover, KLD (Kullback-Leibler Divergence) and MSE (Mean Square Error) lead to gradient decay \cite{bib22} when the student model converges on synthetic images. It will deactivate the generator learning and make the minimax game fails. To tackle this problem, Mean Absolute Error (MAE) is adopted to compute the discrepancy estimation loss:

\begin{equation}
	L_{DE}=E_{z\sim p_z(z)}\left[\frac{1}{n}\left\Vert T\left(G\left(z\mid y\right)\right)-S\left(G\left(z\mid y\right)\right)\right\Vert_1\right] \label{eq5}
\end{equation}
where $T\left(G\left(z\mid y\right)\right)$ and $S\left(G\left(z\mid y\right)\right)$ are the outputs of teacher and student networks. The student and teacher models should produce the same predictions for synthetic images to minimize $L_{DE}$ in the training process. However, this function will induce the generator to craft easy samples. With these easy samples, the student network will quickly learn to produce the same prediction as the teacher model, but it only learns shallow knowledge. Actually, deep latent knowledge can be transferred by exploring hard samples: we conversely maximize this loss to move the generator towards the areas where the current student network is not well trained. 

Combining the loss functions mentioned in Eq.\ref{eq1}, \ref{eq4} and \ref{eq5}, we define the generative loss as:
\begin{equation}
	L_G=-L_{DE}+\lambda_{US}L_{US}+\lambda_{CM}L_{\textsc{CM}} \label{eq6}
\end{equation}
where $\lambda_{US}$ and $\lambda_{CM}$ are scale weights to trade off different loss terms. In the generation stage, we train the proposed conditional generator to craft training samples by minimizing the generation loss. Besides, we utilize batch-norm statistics to promote the training process further.

\subsection{Distillation based on Generated Images}

In section \ref{sec3.1}, discrepancy estimation loss is employed to measure outputs discrepancy of the teacher and student networks. We maximize this loss and induce the generator to produce hard samples. In the distillation process, we reversely minimize this loss function by forcing the student network to imitate teacher output. Thus, we construct an adversarial learning framework. The generator consistently produces hard samples and enables the student network to explore unseen regions.

Besides estimating discrepancy, ground truth is also beneficial for distillation and is commonly used in data-driven KD. Although ground truth in data-free settings is unavailable. We utilize the information from preset labels in Eq.\ref{eq1} to facilitate distillation and narrow the data-free and data-driven KD gap. The ground truth loss is defined as:
\begin{equation}
	L_{GT}=\frac{1}{n}\sum_{i=1}^{n}H\left({\rm softmax} \left(l_S^i\right),y^i\right) \label{eq7}
\end{equation}
where $l_S^i=S\left(G\left(z^i\mid y^i\right)\right) $ is the logits of the student network. Different from $ L_{CM} $, we calculate the cross-entropy loss between student predictions and the preset labels. In this situation, the student model has a clear optimization objective.

Although $ L_{DE} $ and $ L_{GT} $ provide helpful information, they are insufficient for complex tasks. The student model cannot learn valuable knowledge if there is a huge capacity gap between the teacher and student networks, especially in data-free settings. Fortunately, intermediate features offer crucial cues to the learning process. In the natural world, attention plays a critical role in human visual experience. Zagoruyko \cite{bib11} claimed that attention maps in CNNs show the area where the network is concerned. In other words, the pixel with a larger attention value has a higher probability of belonging to the object. We extract the feature knowledge hidden in intermediate layers by introducing the attention transfer loss and formulate the activation-based attention as:
\begin{equation}
	F(A)=\sum_{i=1}^{C}\left\Vert A_i \right\Vert ^2 \label{eq8}
\end{equation}
where $A$ denotes activation in each channel, and $C$ is the number of channels. Power and absolute value operations are element-wise. Then the attention transfer loss can be computed as follows:
\begin{equation}
	L_{AT}=\sum_{j\in I} \left \Vert \frac{Q_S^j}{ \Vert Q_S^j \Vert _2} -\frac{Q_T^j}{\Vert Q_T^j \Vert _2} \right \Vert _2 \label{eq9}
\end{equation}
in which $Q_S^j=vec\left(F\left(A_S^j\right)\right)$ and $Q_T^j=vec\left(F\left(A_T^j\right)\right)$ are the $j$-th pair attention maps of the student and teacher networks. $I$ is the number of layers that perform attention transfer operations, and $vec(\cdot)$ converts the feature maps into vector form. We guide the student to mimic teacher attention maps by minimizing $ L_{AT} $.

Combining the loss functions in Eq.\ref{eq5}, \ref{eq7} and \ref{eq9}, we yield the distillation loss:
\begin{equation}
	L_{KD}=L_{DE}+\lambda_{GT}L_{GT}+\lambda_{AT}L_{AT} \label{eq10}
\end{equation}
where $\lambda_{GT}$ and $\lambda_{AT}$ are hyper-parameters to balance different components.

Based on the distillation loss $ L_{KD} $, the student network learns to extract features and make predictions. It is noted that previous work did not fully use teacher knowledge, consequently leading to an accuracy gap. In contrast, our method gets better performance through efficient information extraction.

\subsection{Conditional Generative Data-free Knowledge Distillation Framework}

\begin{algorithm}
	\caption{Conditional Generative Data-free Knowledge Distillation Framework}\label{algo1}
	\begin{algorithmic}[1]
		\Require A given pre-trained teacher network $T$, a normal distribution $p_z$ and hyper-parameters for balancing different terms
		\Ensure Student network $S$ and generator $G$
		\State Initialize the parameters in student network $S$ and generator $G$
		\While{not converged}
		\State \textbf{Train the Generator}
		\State Random sample a set of noise vectors $\left\{z^i\right\}_{i=1}^n$ from $p_z$ and preset labels $\left\{y^i\right\}_{i=1}^n$;
		\State Generate training samples $x\to G(z\mid y)$;
		\State Calculate the loss of generator $L_G$ in Eq.\ref{eq6};
		\State Update the parameters in generator $G$ according to the gradient;
		\State \textbf{Train the Student Network}
		\For {$k$ iterations}
		\State Sample noise vectors $\left\{z\right\}$ and preset labels$\left\{y\right\}$to craft training samples $x$;
		\State Compute distillation loss $L_{KD}$ in Eq.\ref{eq10};
		\State Update the student network $S$ using back-propagation;
		\EndFor
		\EndWhile
	\end{algorithmic}
\end{algorithm}

The conditional generative data-free knowledge distillation framework for learning efficient portable networks is summarized in Algorithm \ref{algo1}. The training procedure is divided into the generation and distillation stages. We alternately update the parameters in student $S$ and generator $G$, while the teacher model $T$ is fixed.

In the generation stage, we sample a set of preset labels $\left\{y^i\right\}_{i=1}^n$ and noises $\left\{z^i\right\}_{i=1}^n$ to craft training images $G(z\mid y)$. These images will have the same label distribution as the source dataset. $G$ learns to produce meaningful samples with a similar distribution to the original data by minimizing $ L_{G} $ in Eq.\ref{eq6}. Since minimizing $ L_{G} $ will increase $ L_{DE} $, the generator tends to produce hard samples. In the distillation stage, the generated images are employed as training data and force $S$ to explore unseen data space. By minimizing $ L_{KD} $ in Eq.\ref{eq10}, we guide $S$ to learn the knowledge from $T$. The student network not only learns teacher output but also mimics attention maps. Unlike the generator training, we update the student network $k$ times each epoch to ensure it converges.

Due to $ L_{DE} $ decreasing with $ L_{KD} $ while increasing with $ L_{G} $, we construct an adversarial process between the generation and distillation process. 
In the whole adversarial training process, $G$ continuously produces hard samples, and $S$ consistently learns valuable knowledge from $T$. We repeat the two training processes alternately until both models converge.

\section{Experiments}\label{sec4}
In this section, we conduct extensive experiments on different datasets to validate the effectiveness of the proposed method. Furthermore, a series of ablation experiments are performed to analyze different components in this work.

\subsection{Datasets and Models}
We conduct experiments on several datasets, including MNIST, CIFAR, Caltech101, and part of ImageNet.

\noindent
\textbf{MNIST.} 
MNIST \cite{bib54} is a well-known image dataset of handwritten digits composed of 28$\times$28 pixels gray-level images. It is very simple and only contains ten classes (from 0 to 9) with 60,000 training and 10,000 testing images. We adopt LeNet-5 \cite{bib55} and LeNet-5-Half \cite{bib21} as the teacher and student networks.

\noindent
\textbf{CIFAR.} 
CIFAR10 and CIFAR100 \cite{bib56} are more complex image datasets than MNIST. CIFAR10 consists of 32$\times$32 pixels color images in 10 classes, while CIFAR100 has 100 classes. They both contain 50,000 training samples and 10,000 testing samples.

\noindent
\textbf{Caltech101.} 
Caltech101 dataset \cite{bib57} consists of 101 classes. Each class has 40 to 800 images. In the training process, we randomly split 80\% images as the training set and 20\% images as the testing set. These images are larger than 128$\times$128 pixels.

\noindent
\textbf{ImageNet.} ImageNet1k\cite{bib58} dataset provides more than 1.28 million training and 50,000 validation samples in 1000 classes. Due to the complexity of the dataset and the limitation of computational resources, we decompose this task into several subtasks on different datasets: ImageNet100 and ImgaeNet32. 
ImageNet100 consists of 100 randomly selected classes from the original ImageNet dataset. ImageNet32 is a down-sample version of ImageNet. It includes all images of ImageNet, resized to 32$\times$32 pixels.

ResNet34 and ResNet18 \cite{bib59} are adopted as the teacher and student networks for all datasets except MNIST.

\subsection{Evaluation Metrics and Talent Selection Strategy}

\subsubsection{Evaluation Metrics}
Prediction accuracy is often used to evaluate different classification algorithms, while the accuracy gap between teacher and student models is employed in distillation works. However, teacher models in each work have different accuracy, and it is not easy to compare the effectiveness of different distillation methods. In this paper, we propose an additional evaluation metric and define it as relative accuracy:
\begin{equation}
	Rel_{acc}=\frac{S_{acc}}{T_{acc}}\times 100\% \label{eq11}
\end{equation}
where $T_{acc}$ and $S_{acc}$ are the accuracies of teacher and student networks. With relative accuracy, we can directly compare the effectiveness of different distillation methods.

\subsubsection{Talent Selection Strategy}
If the effectiveness of methods or hyper-parameters has not been confirmed, completing training will waste resources. We propose a simple talent selection strategy to reduce training time and computational cost. The strategy is a manual tuning technique that helps us verify new methods and adjust hyper-parameters with minimal consumption. For example, if the model requires 200 training epochs and adjusts the learning rates twice, we train it for 50 epochs or less and change the learning rates at the 40th epoch. After several selection experiments, we can verify current settings by comparing the results and trends. Then we choose 2-4 group better settings for complete training. Although the final results may not be globally optimal, it is a balance between network performance and computational cost.
\subsection{Experimental results}
We implement all experiments with PyTorch and do not use any data augmentation technique for a fair comparison. For all datasets, SGD with momentum 0.9 and weight decay $5\times 10^{-4}$ is applied to update the student networks, and Adam is employed to optimize the generators. We use the generators following \cite{bib23} and make adjustments to match our method.
\subsubsection{Experiments on MNIST}
We first implement the method on MNIST. The images are resized to 32$ \times $32 pixels in the training process, and the batch size is 512. The initial learning rates are 0.01 and 0.001 for student and generator, respectively. For MNIST, we train the models 60 epochs and decay learning rates at the 50th epoch. Each epoch iterates 50 steps, $k$ equals 5. In this experiment, the student network will quickly converge to high accuracy. For simple tasks such as MNIST, we should focus on the training epoch and avoid the model over-fitting.

Table\ref{tab1} reports the results of different data-free distillation methods on MNIST. With the absence of real training data, DAFL \cite{bib22}, DFAD \cite{bib23} and DFKD \cite{bib45} obtained 99.28\%, 99.32\% and 99.53\% relative accuracy, respectively. After completing adversarial training, the student network in our method achieves better results of 98.62\% prediction accuracy and 99.65\% relative accuracy. We get a little but non-negligible gain when accuracy is very close to the upper bound. It demonstrates that our method can explore larger data space.

\begin{table}
	\begin{center}
		\centering
		\caption{Prediction accuracy and relative accuracy of different methods on MNIST dataset. Teacher models have different accuracy in different works.} \label{tab1}
		\begin{tabular*}{1.0 \textwidth}{@{\extracolsep{\fill}}cccc@{\extracolsep{\fill}}}
			\hline
            \hline
			\textbf{Method} & $ T_{acc} $(\%) &  $ S_{acc} $(\%) &  $ Rel_{acc} $(\%)\\
			\hline
			DAFL \cite{bib22}    & 98.91  	 & 98.20 		 & 99.28  \\
			DFAD \cite{bib23}    & 98.97   	 & 98.30  	 	 & 99.32  \\
			DFKD \cite{bib45}    & 98.97  	 & 98.45 		 & 99.53  \\
			Ours 	 & 98.97   	 & \textbf{98.62}		 & \textbf{99.65}  \\
			\hline
            \hline
			\end{tabular*}
	\end{center}
\end{table}

\subsubsection{Experiments on CIFAR} \label{sec4.3.2}
For CIFAR datasets, the initial learning rate of student networks and generators are 0.1 and 0.001, respectively. We use the training epoch of 250 and decay the learning rates twice. In each epoch, we iterate the training process 100 steps and use the batch size of 256. Under this setting, the student model sees 256$\times$100$\times$5 (k=5) synthetic images in each epoch. It is even more than the images in the original CIFAR datasets. In this case, student models can sufficiently explore the data space and get good results. Since CIFAR is more complex than MNIST and will take a lot of training time, we adopt talent selection strategy to tune the hyper-parameters. After the selection process, three better settings were chosen for complete training.

\begin{table}
	\begin{center}
		\caption{Experimental results of different methods on CIFAR datasets. Teacher models have different accuracy in different works. We achieve a comparable result on CIFAR10 and state-of-the-art on CIFAR100.}\label{tab2}%
		\begin{tabular*}{1.0 \textwidth}{@{\extracolsep{\fill}}lllllll@{\extracolsep{\fill}}}
			\hline 
            \hline
			\multirow{2}{*}{\textbf{Method}} & \multicolumn{3}{c}{\textbf{CIFAR10}} & \multicolumn{3}{c}{\textbf{CIFAR100}} \\
			\cline{2-4}\cline{5-7}%
			  & $ T_{acc} $(\%) & $ S_{acc} $(\%) & $ Rel_{acc} $(\%) & $ T_{acc} $(\%) & $ S_{acc} $(\%) & $ Rel_{acc} $(\%)\\
			\hline
			DAFL \cite{bib22} & 95.58  	 & 92.22 	& 96.48    & 77.84    & 74.47    & 95.67  \\
			DFAD \cite{bib23} & 95.54  	 & 93.30 	& 97.65    & 77.50    & 67.70    & 87.35  \\
			LS-GDFD \cite{bib24}  & 95.05  	 & \textbf{95.02} 	& \textbf{99.97}    & 77.26    & 76.42    & 98.91  \\
			CMI \cite{bib46}  & 95.70	& 94.84 	& 99.10    & 78.05    & 77.04    & 98.97  \\
			DDAD \cite{bib60}  & 95.54	& 94.81 	& 99.23    & 77.50    & 75.04    & 96.83  \\
			DFKD \cite{bib45} & 95.32	& 92.84 	& 97.40    & 77.28    & 74.45    & 96.34  \\
			Ours  & 95.54	& \textbf{95.19}   & \textbf{99.63}  & 77.52    & \textbf{76.80}    & \textbf{99.07}  \\
			\hline
            \hline
		\end{tabular*}
	\end{center}
\end{table}

Table \ref{tab2} summarizes the experimental results of different works on CIFAR. For CIFAR10, Google presented LS-GDFD \cite{bib24} and achieved the best results of 99.97\% relative accuracy. It only has a 0.03\% prediction accuracy decay. We achieve comparable results despite not having as many computational resources as Google (they used the batch size of 32678, and ours is 256). The trained student network in our paper yields 99.63\% relative accuracy and 95.19\% prediction accuracy on CIFAR10. It is only slightly worse than LS-GDFD and far outperformed other methods. For CIFAR100, which is more complex than CIFAR10, we obtain the best results of 76.80\% student accuracy and 99.07\% relative accuracy. It shows that our method is not only applicable for simple tasks but also for complex datasets. It is worth noting that LS-GDFD trains one generator for each category of images to avoid mode collapse and improve results. We only compare the results with those obtained on a single generator for a fair comparison. 

\begin{figure}
	\centering
	\includegraphics[width=1.0\textwidth]{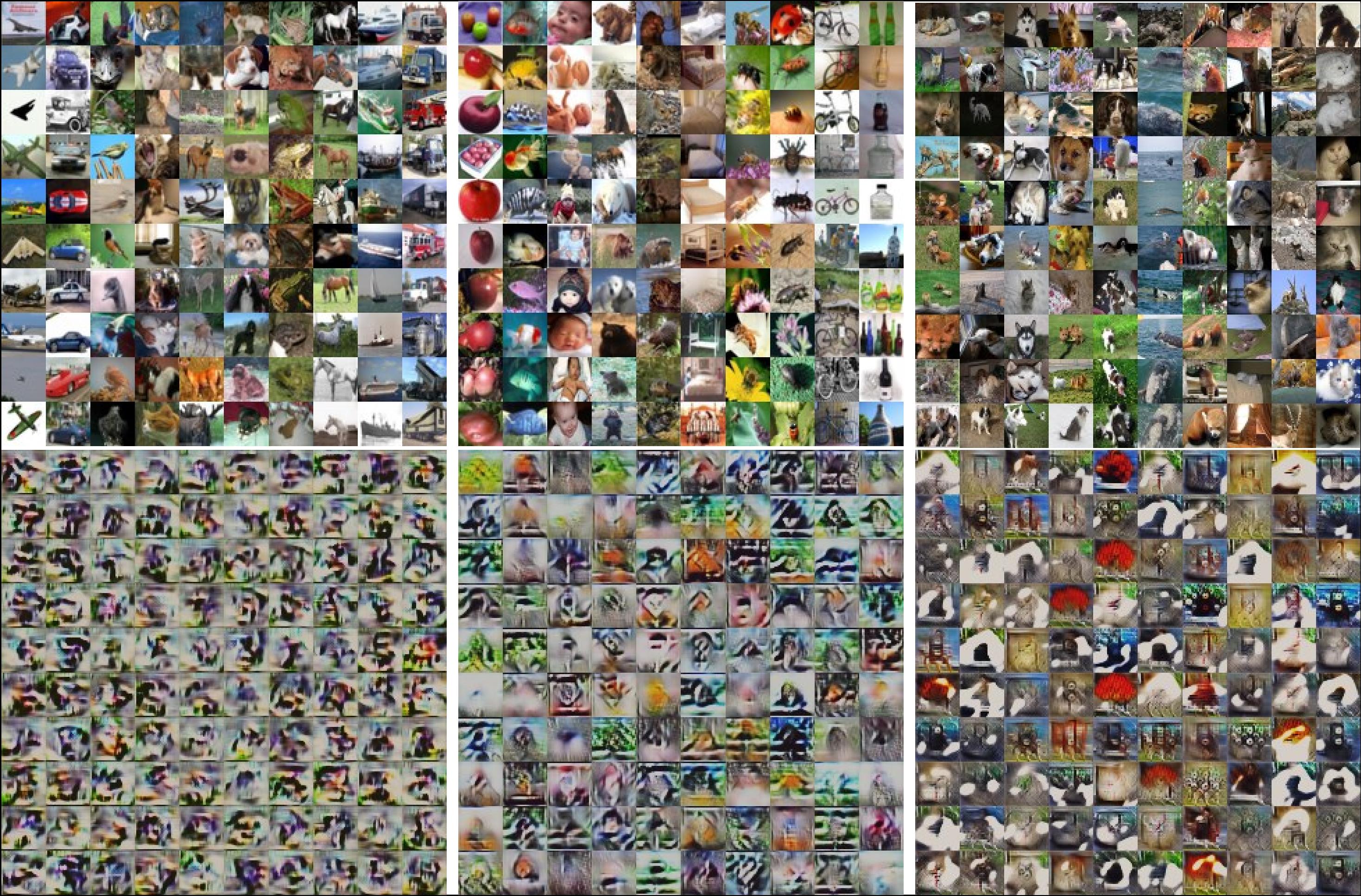}
	\caption{ Synthetic images of CIFAR10 (left), CIFAR100 (middle) and ImageNet32(right), each column has the same category. Top: real images sampled from real datasets. Bottom: images generated by the conditional generator. Notably, the generator tends to produce semantic rather than natural images because there are no real images as the learning target.}\label{fig2}
\end{figure}

Fig.\ref{fig2} (left and middle) shows the synthetic images on CIFAR datasets. We can find that data-free generators tend to produce semantic images due to the lack of real images, while the conventional GAN produces realistic images. 

Furthermore, we multiply the random noise and label embedding rather than concatenate them. We find that the generator will easily produce images with the prominent class feature when adopting the concatenate operation. However, it does not achieve good results. In contrast, the generator with the multiply operation tends to produce weird images but leads to better distillation performance. We think the multiply operation will craft hard samples, while hard samples will induce the student model to explore unseen data space and get better performance.

\subsubsection{Experiments on Caltech101}
We extend experiments to other high-resolution image data to further validate our method. In the training process, we resize the images 128$\times$128. Similar to the experiments on CIFAR, the talent selection strategy is applied to Caltech101. In order to generate larger images, we choose the DCGAN \cite{bib61} and make some adjustments to match our method. The generator in DCGAN is generally used in data-free distillation works and will provide a fair comparison.

\begin{figure}[ht]
	\centering
	\includegraphics[width=1.0 \textwidth]{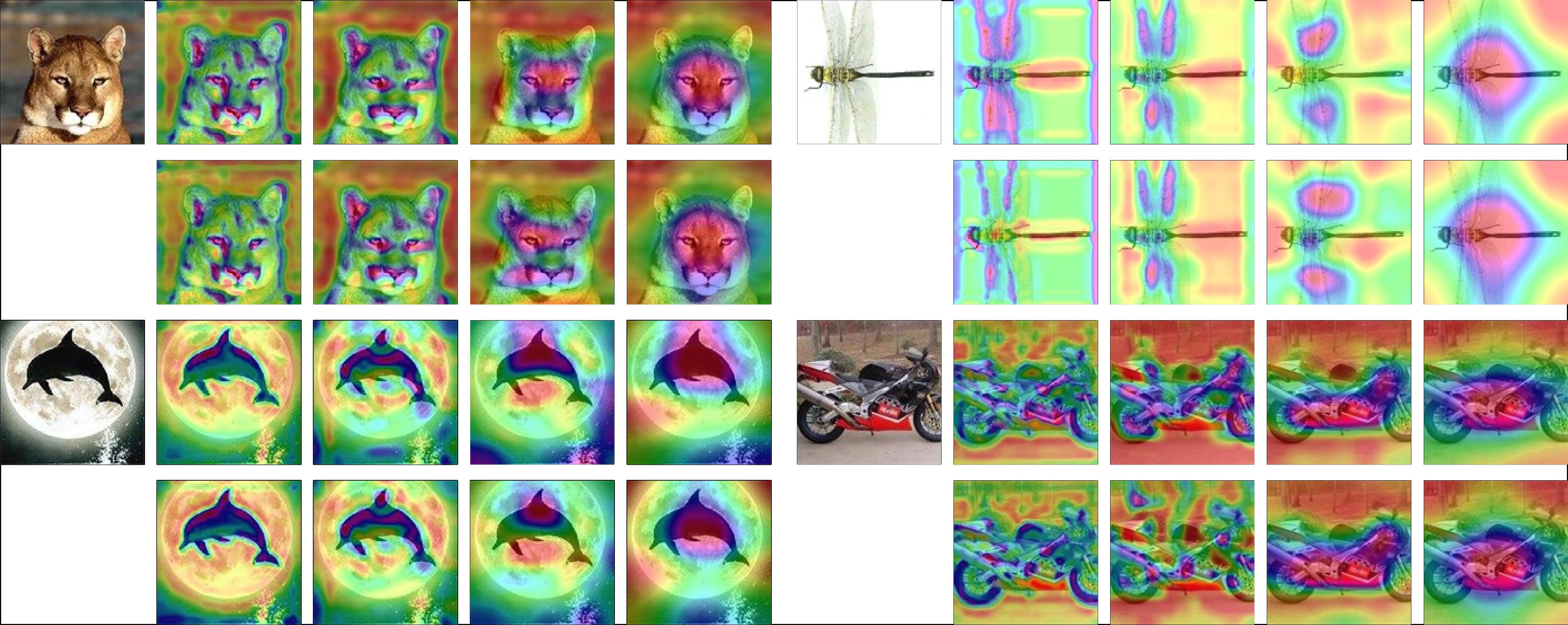}
	\caption{Score-weighted class activation heatmaps on images \cite{bib62}. The top rows are teacher heatmaps, and the bottom rows are student heatmaps. The images from left to right show the features extracted from shallow to top layers of the networks. Shallow filters extract local features while top layers reflect full objects. Best viewed on screen.}\label{fig3}
\end{figure}

Table \ref{tab3} reports the experimental results of Caltech101. While DFAD and DDAD obtained 73.5\% and 75.01\% prediction accuracy, our method gets 76.46\% student accuracy and 99.84\% relative accuracy. We get state-of-the-art results and a 1.9\% relative accuracy gain. It shows that our method can be extended to tasks with larger images. To validate the efficacy of feature information, we visualize the attention maps of teacher and student models in Fig.\ref{fig3}. Different colors indicate how much attention it receives, while the magenta region receives the most attention. The visualization results show the student model can well notice the local and global features of images. In other words, the student has well learned the knowledge from the teacher model.

\begin{table}
	\begin{center}
		\centering
		\caption{Prediction accuracy and relative accuracy on Caltech101. We obtain state-of-the-art results.}\label{tab3}%
		\begin{tabular*}{1.0 \textwidth}{@{\extracolsep{\fill}}llll@{\extracolsep{\fill}}}
			\hline
            \hline
			\textbf{Method} & $ T_{acc} $(\%) & $ S_{acc} $(\%) & $ Rel_{acc} $(\%)\\
			\hline
			DFAD \cite{bib23} & 76.60   	 & 73.50  	 	 & 95.95  \\
			DDAD \cite{bib60} & 76.60  	 & 75.01 		 & 97.92  \\
			Ours 	& 76.58   	 & \textbf{76.46}		 & \textbf{99.84}  \\
			\hline
            \hline
		\end{tabular*}
	\end{center}
\end{table}

\subsubsection{Experiments on ImageNet}
ImageNet is the most well-known image classification dataset, providing 1000 classes of images with a resolution greater than 224. However, as far as we know, only LS-GDFD implements data-free distillation on this task. Based on the ample computing resources of Google, they train 1000 generators for each class. It is helpful but requires heavy computational resources. To this end, we decompose this task into several subtasks on two datasets in this work. Table \ref{tab4} summarises the results on ImageNet32 and ImagNet100. 

For ImageNet32, we validate the distillation performance for more classes. The images in the ImageNet dataset are more diverse than in other datasets. Train one generator for all classes will inevitably meet the model collapse. Even so, our method gets 50.48\% relative accuracy with only one generator, which exceeds the results of LS-FDGD obtained on 100 generators. Our method will perform better under the same settings since multi-generators can efficiently avoid model collapse. Fig.\ref{fig2} (right) shows the generated images of ImageNet32. Then we randomly sample 100 classes from the original ImageNet dataset to make up the ImageNet100 dataset and conduct experiments. We resize the images to 224$ \times $224 and verify the efficacy of generating larger images. For complex generation tasks (e.g., the images in original ImageNet are more diversity), some special designs ensure efficient image generation (e.g., BigGAN \cite{bib63} and SAGAN \cite{bib64}). These designs are beneficial for this work but lead to unfair comparison. Some previous attempts \cite{bib65,bib66} claimed that high data diversity worsens image generation. While not using any special designs, we achieve an acceptable result of 71\% relative accuracy on the size of 224 $\times$ 224. Moreover, the distillation performance will get degraded when the capacity gap between the teacher and student models is significant \cite{bib37}. In the future, we will explore this area to reduce performance degradation.

In the above sections, we perform extensive experiments on different tasks with image sizes ranging from 32 to 224 and image classes ranging from 10 to 1000. The experimental results demonstrate that our method can be well applied to most datasets.
\begin{table}
	\begin{center}
		\centering
		\caption{Distillation results on different versions of ImageNet datasets. $ g $ represents the number of generators. The teacher and student networks are ResNet34 and ResNet18.}\label{tab4}%
		\begin{tabular*}{1.0 \textwidth}{@{\extracolsep{\fill}}lllll@{\extracolsep{\fill}}}
			\hline
            \hline
			\textbf{Datasets} &\textbf{Method} & $ T_{acc} $(\%) & $ S_{acc} $(\%) & $ Rel_{acc} $(\%)\\
			\hline
			\multirow{3}{*}{ImgaeNet32} & \multirow{2}{*}{LS-GDFD \cite{bib24}} & \multirow{2}{*}{59.68}   	 & 15.85($ g=1 $)  	 	 & 25.56  \\
			\multirow{3}{*} & \multirow{2}{*} & \multirow{2}{*} 	 & 29.40($ g=100 $) 		 & 49.26  \\
			\cline{2-5} & Ours 	& 50.04   	& \textbf{25.26}		& \textbf{50.48}  \\
			\hline
            ImageNet100 & Ours & 90.02 & \textbf{64.06} & \textbf{71.16}\\
			\hline
            \hline
		\end{tabular*}
	\end{center}
\end{table}

\subsection{Ablation Study}
In the above sections, we have conducted extensive experiments on different datasets and demonstrated the effectiveness of our method. However, our framework contains several loss components that play different roles. We will analyze the following novel loss functions, including ground truth loss ($ L_{GT} $), class matching loss ($ L_{CM} $) and attention transfer loss ($ L_{AT} $). In this work, we make the preset labels supervise the student model by utilizing $ L_{GT} $, and it only works when we use $ L_{CM} $ to control the image class. To explore these functions, we perform a series of ablation experiments on CIFAR100 following the experimental settings in section \ref{sec4.3.2}. The results of various design components are summarized in Table \ref{tab5}.

Student accuracy without any novel loss function is 74.81\%. We first introduce the attention transfer loss to this framework, improving the student accuracy to 76.34\%. We get a 1.53\% performance gain on CIFAR100. It suggests that the attention transfer can remarkably improve distillation performance. Actually, a trained teacher model will focus on the target area of input images and capture valuable information. With the guidance of attention transfer, student network will learn the knowledge from teacher and achieve better performance.

Many feature-based distillation works presented efficient information extraction methods except for transferring the attention maps. Although these data-driven methods improved distillation performance, the challenges remain in data-free tasks. In practice, we can represent hidden information in many forms, and the student model can well learn the knowledge by exploring training data. However, the knowledge with some forms may not be easy to learn, especially when the training data is not accessible. To this end, more effective knowledge representation methods should be explored to attain efficient knowledge transfer.

In the following experiments, we find $ L_{CM} $ also prominently improves student accuracy and gets a 0.7\% accuracy gain. It is because the generator will converge better when subjected to additional constraints. We obtain 76.71\% prediction accuracy when combining $ L_{CM} $ with $ L_{AT} $ and 75.56\% accuracy when combining $ L_{CM} $ with $ L_{GT} $. Although the ground truth loss does not lead to remarkable improvements, it greatly accelerates the model converges. 
When using all of these components, our method gets the best performance of 76.80\%. The ablation experiments demonstrate the effectiveness of each component in our work. In addition, the generator produces meaningless images in the early stage, so the unsupervised loss $ L_{UN}$ is necessary for stable training.
\begin{table}
	\begin{center}
		\centering
		\caption{Ablation Study by cutting different components. $ L_{CM} $, $ L_{GT} $, and $ L_{AT} $ represent the class matching loss, ground truth loss, and attention transfer loss, respectively.}\label{tab5}%
		\begin{tabular*}{1.0\textwidth}{@{\extracolsep{\fill}}ccccccc@{\extracolsep{\fill}}}
			\hline
            \hline
			$ L_{CM} $	&  		& \checkmark &  		  & \checkmark & \checkmark  & \checkmark \\
			$ L_{GT} $ 		&  		&  			&  			  & \checkmark &  			 & \checkmark \\
			$ L_{AT} $	&  		&  			&  \checkmark &  		   &  \checkmark & \checkmark \\
			\hline
			$ S_{acc} $(\%)		& 74.81	& 75.52		& 76.34		  & 75.56	   & 76.71		 & \textbf{76.80} 	  \\
			\hline
            \hline
		\end{tabular*}
	\end{center} 
\end{table}

\section{Conclusion}\label{sec5}
In this work, we present a conditional generative data-free distillation framework. It trains lightweight student networks through efficient image generation and knowledge distillation. This framework can be divided into two adversarial processes. We treat preset labels as ground truth in the generation process to train a conditional generator. Semi-supervised training can avoid the class-imbalanced issue in heterogeneity tasks through label sampling. In the distillation process, the generator produces hard samples as training data. We force the student model to fit these hard samples, thus learning teacher knowledge and exploring unseen data space. In conclusion, we construct a whole adversarial learning framework base on the data generation and exploring processes. Experimental results show that the proposed method remarkably improves student performance and yields state-of-the-art results on most datasets. Furthermore, this method can be well applied to extended data-free tasks, such as federated learning on medical images.\\
\\
\noindent \textbf{CRediT authorship contribution statement}\\

\indent \textbf{Xinyi Yu:}\ Conceptualization, Formal analysis.
\textbf{Ling Yan:}\ Conceptualization, Methodology, Validation, Writing original draft.
\textbf{Yang Yang:}\ Investigation, Visualization, Writing original draft.
\textbf{Libo Zhou:}\ Investigation, Data curation.
\textbf{Linlin Ou:}\ Writing-review \& editing, Supervision.
\\

\noindent \textbf{Declaration of Competing Interest}\\
\\
\indent The authors declare that they have no known competing financial interests or personal relationships that could have appeared to influence the work reported in this paper.\\

\noindent \textbf{Acknowledgments}\\
\\
\indent This work was supported by National Key R\&D Program of China (Grant No.2018YFB1308400) and Natural Science Foundation of Zhejiang Province (Grant No.LY21F030018 and No.LQ22F030021)\par

\bibliographystyle{elsarticle-num}
\bibliography{CGDD}

\end{document}